\documentclass[12pt]{article}
\usepackage[T1]{fontenc}

\usepackage{tgtermes}
\usepackage{amsmath}
\usepackage{amsfonts}
\usepackage{scalefnt,letltxmacro}
\LetLtxMacro{\oldtextsc}{\textsc}
\renewcommand{\textsc}[1]{\oldtextsc{\scalefont{1.10}#1}}

\usepackage[bitstream-charter]{mathdesign}

\usepackage[usenames,dvipsnames]{xcolor}
\definecolor{shadecolor}{gray}{0.9}

\usepackage[final,expansion=alltext]{microtype}
\usepackage[english]{babel}
\usepackage[parfill]{parskip}
\usepackage{afterpage}
\usepackage{framed}
\usepackage{nicefrac}

\usepackage{lineno}

\usepackage{ragged2e}

\DeclareRobustCommand{\parhead}[1]{\vspace{0.05in} \textbf{#1} }

\usepackage{graphicx}
\usepackage[labelfont=bf]{caption}
\usepackage{subfig}

\usepackage{booktabs}
\usepackage{tabu}

\usepackage{natbib}

\usepackage[algoruled]{algorithm2e}
\usepackage{listings}
\usepackage{fancyvrb}
\fvset{fontsize=\normalsize}

\usepackage[colorlinks,linktoc=all]{hyperref}
\usepackage[all]{hypcap}
\hypersetup{citecolor=Violet}
\hypersetup{linkcolor=black}
\hypersetup{urlcolor=MidnightBlue}

\usepackage
[acronym,smallcaps,nowarn,section,nonumberlist]{glossaries}
\glsdisablehyper{}

\usepackage{listings}
\lstdefinestyle{alp_style}{
    commentstyle=\color{OliveGreen},
    numberstyle=\tiny\color{black!60},
    stringstyle=\color{BrickRed},
    basicstyle=\ttfamily\scriptsize,
    breakatwhitespace=false,
    breaklines=true,
    captionpos=b,
    keepspaces=true,
    numbers=none,
    numbersep=5pt,
    showspaces=false,
    showstringspaces=false,
    showtabs=false,
    tabsize=2
}
\newcommand{\expfam}{\textrm{ExpFam}}

\DeclareRobustCommand{\mb}[1]{\ensuremath{\boldsymbol{\mathbf{#1}}}}

\newcommand{\mbx}{\mb{x}}

\newcommand{\mbalpha}{\mb{\alpha}}

\newcommand{\mbrho}{\mb{\rho}}

\newcommand{\g}{\, | \,}
 
\newacronym{ELBO}{elbo}{evidence lower bound}
\newacronym{GMM}{gmm}{Gaussian mixture model}
\newacronym{KL}{kl}{Kullback-Leibler}
\newacronym{LDA}{lda}{latent Dirichlet allocation}
\newacronym{SGD}{sgd}{stochastic gradient descent}
\newacronym{SVI}{svi}{stochastic variational inference}

\newacronym{GLM}{glm}{generalized linear model}
\newacronym{CBOW}{cbow}{continuous bag of words}
\newacronym{PCA}{pca}{principal component analysis}
\newacronym{NPLM}{nplm}{neural probabilistic language model}
\newacronym{vLBL}{vlbl}{vector log-bilinear language model}
\newacronym{ivLBL}{ivlbl}{inverse vector log-bilinear language model}

\newacronym{efemb}{ef-emb}{exponential family embedding}
\newacronym{gemb}{g-emb}{Gaussian embedding}
\newacronym{ngemb}{ng-emb}{nonnegative Gaussian embedding}
\newacronym{pemb}{p-emb}{Poisson embedding}
\newacronym{apemb}{ap-emb}{additive Poisson embedding}
\newacronym{bemb}{b-emb}{Bernoulli embedding}
\newacronym{FA}{fa}{factor analysis}
\newacronym{HPF}{hpf}{hierarchical Poisson factorization}
\newacronym{NMF}{nmf}{nonnegative matrix factorization}
\newacronym{KNN}{knn}{K-nearest neighbors}

\usepackage[nonatbib,final]{nips_2016}

\title{Exponential Family Embeddings}

\author{
  Maja Rudolph\\
  Columbia University\\
  \And
  Francisco J.\ R.\ Ruiz\\
  Univ.\ of Cambridge \\
  Columbia University\\ 
  \And
  Stephan Mandt\\
  Columbia University\\
  \And
  David M.\ Blei\\
  Columbia University
}
\begin{document}

\maketitle

\begin{abstract}
  Word embeddings are a powerful approach for capturing semantic
  similarity among terms in a vocabulary.  In this paper, we develop
  \textit{exponential family embeddings}, a class of methods that
  extends the idea of word embeddings to other types of
  high-dimensional data. As examples, we studied neural data with
  real-valued observations, count data from a market basket analysis,
  and ratings data from a movie recommendation system.  The main idea
  is to model each observation conditioned on a set of other
  observations.  This set is called the context, and the way the
  context is defined is a modeling choice that depends on the problem.
  In language the context is the surrounding words; in neuroscience
  the context is close-by neurons; in market basket data the context
  is other items in the shopping cart.  Each type of embedding model
  defines the context, the exponential family of conditional
  distributions, and how the latent embedding vectors are shared
  across data. We infer the embeddings with a scalable algorithm based
  on stochastic gradient descent. On all three applications---neural
  activity of zebrafish, users' shopping behavior, and movie
  ratings---we found exponential family embedding models to be more
  effective than other types of dimension reduction.  They better
  reconstruct held-out data and find interesting qualitative
  structure.
\end{abstract}

\section{Introduction}

Word embeddings are a powerful approach for analyzing
language~\citep{bengio2006neural,mikolov2013efficient,mikolov2013distributed,pennington2014glove}.
A word embedding method discovers distributed representations of
words; these representations capture the semantic similarity between
the words and reflect a variety of other linguistic
regularities~\citep{rumelhart1986learning,bengio2006neural,mikolov2013linguistic}.
Fitted word embeddings can help us understand the structure of
language and are useful for downstream tasks based on text.

There are many variants, adaptations, and extensions of word
embeddings
\citep{mikolov2013efficient,mikolov2013distributed,mnih2013learning,levy2014neural,pennington2014glove,vilnis2014word},
but each reflects the same main ideas.  Each term in a vocabulary is
associated with two latent vectors, an \textit{embedding} and a
\textit{context vector}.  These two types of vectors govern
conditional probabilities that relate each word to its surrounding context.
Specifically, the conditional
probability of a word combines its embedding and the context vectors
of its surrounding words.  (Different methods combine them
differently.)  Given a corpus, we fit the embeddings 
by maximizing the conditional probabilities of the observed
text.

In this paper we develop the \textit{\gls{efemb}}, a class of models
that generalizes the spirit of word embeddings to other types of
high-dimensional data. Our motivation is that other types of data can
benefit from the same assumptions that underlie word embeddings,
namely that a data point is governed by the other data in its
context. In language, this is the foundational idea that words with
similar meanings will appear in similar contexts
\citep{harris1954distributional}. We use the tools of exponential
families \citep{brown1986fundamentals} and \glspl{GLM}
\citep{mccullagh1989generalized} to adapt this idea beyond language.

As one example beyond language, we will study
computational neuroscience. Neuroscientists measure sequential neural
activity across many neurons in the brain. Their goal is to discover
patterns in these data with the hope of better understanding the
dynamics and connections among neurons.  In this example, a context
can be defined as the neural activities of other nearby neurons, or as
neural activity in the past.  Thus, it is plausible that the activity
of each neuron depends on its context.  We will use this idea to fit
latent embeddings of neurons, representations of neurons that uncover
hidden features which help suggest their roles in the brain.

Another example we study involves shoppers at the grocery store.
Economists collect shopping data (called ``market basket data'') and
are interested in building models of purchase behavior for downstream
econometric analysis, e.g., to predict demand and market changes. To
build such models, they seek features of items that are predictive of
when they are purchased and in what quantity. Similar to language,
purchasing an item depends on its context, i.e., the other items in
the shopping cart. In market basket data, Poisson embeddings can
capture important econometric concepts, such as items that tend not to
occur together but occur in the same contexts (substitutes) and items
that co-occur, but never one without the other (complements).

We define an \gls{efemb}, such as one for neuroscience or shopping
data, with three ingredients. (1) We define the \textit{context},
which specifies which other data points each observation depends
on. (2) We define the \textit{conditional exponential family}. This
involves setting the appropriate distribution, such as a Gaussian for
real-valued data or a Poisson for count data, and the way to combine
embeddings and context vectors to form its natural parameter.  (3) We
define the \textit{embedding structure}, how embeddings and context
vectors are shared across the conditional distributions of each
observation.  These three ingredients enable a variety of embedding
models.

We describe \gls{efemb} models and develop efficient algorithms for
fitting them. We show how existing methods, such as
\gls{CBOW}~\citep{mikolov2013efficient} and negative
sampling~\citep{mikolov2013distributed}, can each be viewed as an
\gls{efemb}. We study our methods on three different types of
data---neuroscience data, shopping data, and movie ratings data.
Mirroring the success of word embeddings, \gls{efemb} models
outperform traditional dimension reduction, such as exponential family
\gls{PCA}~\citep{collins2001generalization} and Poisson
factorization~\citep{Gopalan2015}, and find interpretable features of
the data.

\parhead{Related work.}  \gls{efemb} models generalize \gls{CBOW}
\citep{mikolov2013efficient} in the same way that exponential family
\gls{PCA} \citep{collins2001generalization} generalizes \gls{PCA},
\glspl{GLM} \citep{mccullagh1989generalized} generalize regression,
and deep exponential families \citep{ranganath2014deep} generalize
sigmoid belief networks \citep{neal1990learning}.  A linear
\gls{efemb} (which we define precisely below) relates to
context-window-based embedding methods such as \gls{CBOW} or the \gls{vLBL}
\citep{mikolov2013efficient,mnih2013learning}, which model a word
given its context. The more general \gls{efemb} relates to embeddings
with a nonlinear component, such as the skip-gram
\citep{mikolov2013efficient} or the \gls{ivLBL} \citep{mnih2013learning}.
(These methods might appear linear but, when viewed as a conditional
probabilistic model, the normalizing constant of each word induces a
nonlinearity.)

Researchers have developed different approximations of the word
embedding objective to scale the procedure.  These include noise
contrastive estimation~\citep{gutmann2010noise,mnih2012fast}, hierarchical
softmax~\citep{mikolov2013distributed}, and negative
sampling~\citep{mikolov2013efficient}.  We explain in
\ref{sec:inference} and \ref{sec:inference_long} how negative sampling corresponds to biased
stochastic gradients of an \gls{efemb} objective.

\section{Exponential Family Embeddings}
\glsresetall

We consider a matrix $\mbx = x_{1:I}$ of $I$ observations, where each $x_i$ is a
$D$-vector. As one example, in language $x_i$ is an indicator vector
for the word at position $i$ and $D$ is the size of the vocabulary. As
another example, in neural data $x_i$ is the neural activity measured at index
pair $i = (n,t)$, where $n$ indexes a neuron and $t$ indexes a time
point; each measurement is a scalar ($D=1$).

The goal of an \gls{efemb} is to derive useful features of the data.
There are three ingredients: a context function, a conditional
exponential family, and an embedding structure. These ingredients work
together to form the objective. First, the \gls{efemb} models each
data point conditional on its context; the context function determines
which other data points are at play. Second, the conditional
distribution is an appropriate exponential family, e.g., a Gaussian
for real-valued data. Its parameter is a function of the embeddings of
both the data point and its context. Finally, the embedding structure
determines which embeddings are used when the $i$th point appears,
either as data or in the context of another point.  The objective is
the sum of the log probabilities of each data point given its context.
We describe each ingredient, followed by the \gls{efemb} objective. Examples are in \ref{sec:examples}.

\parhead{Context.} Each data point $i$ has a {\em context} $c_i$,
which is a set of indices of other data points. The \gls{efemb} models
the conditional distribution of $x_i$ given the data points in its
context.

The context is a modeling choice; different applications will require
different types of context. In language, the data point is a word and
the context is the set of words in a window around it. In neural data,
the data point is the activity of a neuron at a time point and the
context is the activity of its surrounding neurons at the same time
point. (It can also include neurons at future time or in the past.) In
shopping data, the data point is a purchase and the context is the
other items in the cart.

\parhead{Conditional exponential family.} An \gls{efemb} models each
data point $x_i$ conditional on its context $\mbx_{c_i}$. The
distribution is an appropriate exponential family,
\begin{align}
  x_{i} \g \mbx_{c_i} \sim \expfam(\eta_{i}(\mbx_{c_i}), t(x_{i})),
\end{align}
where $\eta_i(\mbx_{c_i})$ is the natural parameter and $t(x_i)$ is
the sufficient statistic. In language modeling, this family is usually
a categorical distribution. Below, we will study Gaussian and Poisson.

We parameterize the conditional with two types of vectors, embeddings
and context vectors. The \textit{embedding} of the $i$th data point
helps govern its distribution; we denote it
$\rho[i] \in \mathbb{R}^{K\times D}$. The \textit{context vector} of
the $i$th data point helps govern the distribution of data for which
$i$ appears in their context; we denote it
$\alpha[i] \in \mathbb{R}^{K\times D}$.

How to define the natural parameter as a function of these vectors is
a modeling choice. It captures how the context interacts with an
embedding to determine the conditional distribution of a data point.
Here we focus on the \textit{linear embedding}, where the natural
parameter is a function of a linear combination of the latent vectors,
\begin{align}
  \label{eqn:linear-embedding}
  \eta_i(\mbx_{c_i}) = f_i\left(\rho[i]^\top \sum_{j \in c_i} \alpha[j] x_j \right).
\end{align}
Following the nomenclature of \glspl{GLM}, we call $f_i(\cdot)$ the
\textit{link function}.  We will see several examples of link
functions in \ref{sec:examples}.

This is the setting of many existing word embedding models, though not
all. Other models, such as the skip-gram, determine the probability
through a ``reverse'' distribution of context words given the data
point. These non-linear embeddings are still
instances of an \gls{efemb}.

\parhead{Embedding structure.} The goal of an \gls{efemb} is to find
embeddings and context vectors that describe features of the data. The
\textit{embedding structure} determines how an \gls{efemb} shares
these vectors across the data. It is through sharing the vectors that
we learn an embedding for the object of primary interest, such as a
vocabulary term, a neuron, or a supermarket product.
In language the same parameters $\rho[i] = \rho$ and
$\alpha[i] = \alpha$ are shared across all positions $i$. In neural
data, observations share parameters when they describe the same
neuron. Recall that the index connects to both a neuron and time point
$i=(n,t)$. We share parameters with $\rho[i] = \rho_n$ and
$\alpha[i] =\alpha_n$ to find embeddings and context vectors that
describe the neurons. Other variants might tie the embedding and
context vectors to find a single set of latent variables,
$\rho[i] = \alpha[i]$.

\parhead{The objective function.} The \gls{efemb} objective sums the
log conditional probabilities of each data point, adding regularizers
for the embeddings and context vectors.\footnote{One might be tempted
  to see this as a probabilistic model that is conditionally
  specified. However, in general it does not have a consistent joint
  distribution \citep{arnold2001conditionally}.}  We use log
probability functions as regularizers, e.g., a Gaussian probability
leads to $\ell_2$ regularization.  We also use regularizers to
constrain the embeddings,e.g., to be non-negative.  Thus, the
objective is
\begin{align}
  \label{eqn:objective}
  \mathcal{L}(\mbrho, \mbalpha)=\sum_{i=1}^I \left( \eta_{i}^\top t(x_{i}) - a(\eta_{i})\right) + \log p(\mbrho) + \log p(\mbalpha).
\end{align}
We maximize this objective with respect to the embeddings and context
vectors.  In \ref{sec:inference} we explain how to fit it with
stochastic gradients.

\ref{eqn:objective} can be seen as a likelihood function for a bank
of \glspl{GLM}~\citep{mccullagh1989generalized}.  Each data point is
modeled as a response conditional on its ``covariates,'' which combine
the context vectors and context, e.g., as in
\ref{eqn:linear-embedding}; the coefficient for each response is the
embedding itself.  We use properties of
exponential families and results around \glspl{GLM} to derive
efficient algorithms for \gls{efemb} models.

\subsection{Examples}
\label{sec:examples}
We highlight the versatility of \gls{efemb} models with three example
models and their variations. We develop the \gls{gemb} for analyzing
real observations from a neuroscience application; we also introduce a
nonnegative version, the \gls{ngemb}. We develop two Poisson embedding
models, \gls{pemb} and \gls{apemb}, for analyzing count data; these
have different link functions. We present a categorical embedding
model that corresponds to the \gls{CBOW} word
embedding~\citep{mikolov2013efficient}. Finally, we present a
\gls{bemb} for binary data.  In \ref{sec:inference} we explain how 
negative sampling~\citep{mikolov2013distributed} corresponds to biased
stochastic gradients of the \gls{bemb} objective.
For convenience, these acronyms are in \ref{tab:acronyms}.

\begin{table}[h!]
  \begin{center}
    \begin{tabular}{ll}
      {\bf \gls{efemb}} & {\it exponential family embedding} \\
      {\bf \gls{gemb}} & {\it Gaussian embedding} \\
      {\bf \gls{ngemb}} & {\it nonnegative Gaussian embedding} \\
      {\bf \gls{pemb}} & {\it Poisson embedding} \\
      {\bf \gls{apemb}} & {\it additive Poisson embedding}\\
      {\bf \gls{bemb}} & {\it Bernoulli embedding}
    \end{tabular}
  \end{center}
  \caption{Acronyms used for exponential family embeddings.}
  \label{tab:acronyms}
\end{table}
\vspace{-7pt}
\parhead{Example 1: Neural data and Gaussian observations.} Consider
the (calcium) expression of a large population of zebrafish
neurons~\citep{ahrens2013whole}. The data are processed to extract the
locations of the $N$ neurons and the neural activity $x_i = x_{(n,t)}$
across location $n$ and time $t$. The goal is to model the similarity
between neurons in terms of their behavior, to embed each neuron in a
latent space such that neurons with similar behavior are close to each
other.

We consider two neurons similar if they behave similarly in the
context of the activity pattern of their surrounding neurons. Thus we
define the context for data index $i =(n,t)$ to be the indices of the
activity of nearby neurons at the same time. We find the \gls{KNN} of
each neuron (using a Ball-tree algorithm) according to their spatial
distance in the brain. We use this set to construct the context
$c_i = c_{(n,t)}=\{(m,t) | m \in \text{\gls{KNN}}(n) \}$.  This
context varies with each neuron, but is constant over time.

With the context defined, each data point $x_i$ is modeled with a
conditional Gaussian. The conditional mean is the inner product from
\ref{eqn:linear-embedding}, where the context is the simultaneous
activity of the nearest neurons and the link function is the identity.
The conditionals of two observations share parameters if they
correspond to the same neuron.  The embedding structure is thus
$\rho[i] = \rho_n$ and $\alpha[i] =\alpha_n$ for all
$i=(n,t)$. Similar to word embeddings, each neuron has two distinct
latent vectors: the neuron embedding $\rho_n\in\mathbb{R}^K$ and the
context vector $\alpha_n\in\mathbb{R}^K$.

These ingredients, along with a regularizer, combine to form a neural
embedding objective. \gls{gemb} uses $\ell_2$ regularization (i.e., a
Gaussian prior); \gls{ngemb} constrains the vectors to be nonnegative
($\ell_2$ regularization on the logarithm. i.e., a log-normal prior).

\parhead{Example 2: Shopping data and Poisson observations.}
We also study data about people shopping. The data contains the
individual purchases of anonymous users in chain grocery and drug
stores. There are $N$ different items and $T$ trips to the stores
among all households. The data is a sparse $N\times T$ matrix of
purchase counts. The entry $x_{i}=x_{(n,t)}$ indicates the number of
units of item $n$ that was purchased on trip $t$. Our goal is to learn
a latent representation for each product that captures the similarity
between them.

We consider items to be similar if they tend to be purchased in with
similar groups of other items. The \textit{context} for observation
$x_{i}$ is thus the other items in the shopping basket on the same
trip. For the purchase count at index $i=(n,t)$, the context is
$c_i = \{ j=(m,t) | m\neq n\}$.

We use conditional Poisson distributions to model
the count data. The sufficient statistic of the Poisson is
$t(x_{i})= x_{i}$, and its natural parameter is the logarithm of the
rate (i.e., the mean). We set the natural parameter as in
\ref{eqn:linear-embedding}, with the link function defined below. The
embedding structure is the same as in \gls{gemb}, 
producing embeddings for the items.

We explore two choices for the link function. \gls{pemb} uses an
identity link function. Since the conditional mean is the
exponentiated natural parameter, this implies that the context items
contribute multiplicatively to the mean. (We use
$\ell_2$-regularization on the embeddings.)  Alternatively, we can
constrain the parameters to be nonnegative and set the link function
$f(\cdot)=\log(\cdot)$.  This is \gls{apemb}, a model with an additive
mean parameterization.  (We use $\ell_2$-regularization 
in log-space.) \gls{apemb} only captures positive
correlations between items.

\parhead{Example 3: Text modeling and categorical observations.}
\glspl{efemb} are inspired by word embeddings, such as \gls{CBOW}
\citep{mikolov2013efficient}. \gls{CBOW} is a special case of an
\gls{efemb}; it is equivalent to a multivariate \gls{efemb} with
categorical conditionals. In the notation here, each $x_i$ is an
indicator vector of the $i$th word. Its dimension is the vocabulary
size. The context of the $i$th word are the other words in a window
around it (of size $w$), $c_i = \{j\neq i | i-w \leq j \leq i+w \}$.

The distribution of $x_i$ is categorical, conditioned on the
surrounding words $\mbx_{c_i}$; this is a softmax regression.  It has
natural parameter as in \ref{eqn:linear-embedding} with an identity
link function. The embedding structure imposes that parameters are
shared across all observed words. The embeddings are shared globally
($\rho[i] = \rho$, $\alpha[i] = \alpha \in \mathbb{R}^{N\times K}$). 
The word and context embedding of the $n^{th}$ word is the 
$n^{th}$ row of $\rho$ and $\alpha$ respectively. 
\gls{CBOW} does not use any regularizer.

\parhead{Example 4: Text modeling and binary observations.}
One way to simplify the \gls{CBOW} objective is with a model of each
entry of the indicator vectors. The data are binary and indexed by
$i=(n,v)$, where $n$ is the position in the text and $v$ indexes the
vocabulary; the variable $x_{n,v}$ is the indicator that word $n$ is
equal to term $v$.  (This model relaxes the constraint that for any
$n$ only one $x_{n,v}$ will be on.)  With this notation, the context
is $c_i = \{(j,v')| \forall v', j\neq n, n-w \leq j \leq n+w \}$; the
embedding structure is $\rho[i] = \rho[(n,v)] = \rho_v$ and
$\alpha[i] = \alpha[(n,v)] =\alpha_v$.

We can consider different conditional distributions in this setting.
As one example, set the conditional distribution to be a Bernoulli
with an identity link; we call this the \gls{bemb} model for text. In
\ref{sec:inference} we show that biased stochastic gradients of the
\gls{bemb} objective recovers negative
sampling~\citep{mikolov2013distributed}. As another example, set the
conditional distribution to Poisson with link $f(\cdot)=\log(\cdot)$.
The corresponding embedding model relates closely to Poisson
approximations of distributed multinomial
regression~\citep{taddy2015distributed}.

\subsection{Inference and Connection to Negative Sampling}
\label{sec:inference}

We fit the embeddings $\rho[i]$ and context vectors $\alpha[i]$ by
maximizing the objective function in \ref{eqn:objective}.  We use
\gls{SGD} with Adagrad~\citep{Duchi2011}. We can derive the analytic gradient of the objective
function using properties of the exponential family (see the Supplement
for details).
The gradients linearly combine the data in summations we can approximate using subsampled minibatches of data. This reduces the computational cost.

When the data is sparse, we can split the gradient into the summation of
two terms: one term corresponding to all data entries $i$ for which
$x_i\neq 0$, and one term corresponding to those data entries $x_i=0$.
We compute the first term of the gradient exactly---when the data is
sparse there are not many summations to make---and we estimate the second term 
by subsampling the zero entries.  Compared to computing
the full gradient, this reduces the complexity when most of the
entries $x_{i}$ are zero.  But it retains the strong information
about the gradient that comes from the non-zero entries.

This relates to negative sampling, which is used to approximate the
skip-gram objective~\citep{mikolov2013distributed}.  Negative sampling
re-defines the skip-gram objective to distinguish target (observed)
words from randomly drawn words, using logistic regression. The
gradient of the stochastic objective is identical to a noisy but
biased estimate of the gradient for a \gls{bemb} model.  To
obtain the equivalence, preserve the terms for the non-zero data and
subsample terms for the zero data.  While an unbiased stochastic
gradient would rescale the subsampled terms, negative sampling does
not.  Thus, negative sampling corresponds to a biased estimate,
which down-weights the contribution of the zeros. See the Supplement
for the mathematical details.

\section{Empirical Study}
\label{sec:experiments}
\glsresetall

We study \gls{efemb} models on real-valued and count-valued data, and
in different application domains---computational neuroscience,
shopping behavior, and movie ratings.  We present quantitative
comparisons to other dimension reduction methods and illustrate how we
can glean qualitative insights from the fitted embeddings.

\subsection{Real Valued Data: Neural Data Analysis}

\parhead{Data.}  We analyze the neural activity of a larval zebrafish,
recorded at single cell resolution for $3000$ time
frames~\citep{ahrens2013whole}.  Through genetic modification,
individual neurons express a calcium indicator when they fire.  The
resulting calcium imaging data is preprocessed by a nonnegative matrix
factorization to identify neurons, their locations, and the
fluorescence activity $x^*_t \in \mathbb{R}^N$ of the individual
neurons over time~\citep{friedrich2015fast}.  Using this method, our
data contains 10,000 neurons (out of a total of 200,000).

We fit all models on the lagged data $x_{t} = x^*_t - x^*_{t-1}$ to
filter out correlations based on calcium decay and preprocessing.%
\footnote{We also analyzed unlagged data but all methods
  resulted in better reconstruction on the lagged data.}  The
calcium levels can be measured with great spatial resolution but the
temporal resolution is poor; the neuronal firing rate is much 
higher than the sampling rate. Hence we ignore all ``temporal structure'' 
in the data and model the simultaneous activity of the neurons. We use 
the \gls{gemb} and \gls{ngemb} from \ref{sec:examples} to model the 
lagged activity of the neurons conditional on the lags of surrounding 
neurons. We study context sizes $c \in \{10,50\}$ and latent dimension
$K \in \{10, 100\}$.

\parhead{Models.} We compare \gls{efemb} to probabilistic \gls{FA},
fitting $K$-dimensional factors for each neuron and $K$-dimensional
factor loadings for each time frame. In \gls{FA}, each entry of the
data matrix is Gaussian distributed, with mean equal to the inner product of the
corresponding factor and factor loading.

\parhead{Evaluation.}  We train each model on a random sample of $90\%$ of the lagged
time frames and hold out $5\%$ each for validation and testing.  With the test
set, we use two types of evaluation.  (1) {\it Leave one out}: For
each neuron $x_{i}$ in the test set, we use the measurements of the
other neurons to form predictions. For \gls{FA} this means the other
neurons are used to recover the factor loadings; for \gls{efemb} this
means the other neurons are used to construct the context.  (2) {\it
  Leave $25 \%$ out}: We randomly split the neurons into 4 folds.
Each neuron is predicted using the three sets of neurons that are out
of its fold.  (This is a more difficult task.)  Note in \gls{efemb},
the missing data might change the size of the context of some neurons.
See \ref{tab:hyper} in \ref{sec:hyper} for the choice of
hyperparameters.

\begin{table}[]
\centering
\small
\begin{tabular}{lllll}
\toprule
                       & \multicolumn{2}{c}{single neuron held out}   & \multicolumn{2}{c}{$25\%$ of neurons held out} \\
  {\bf Model}                & $K=10           $          &$ K=100          $ &$ K=10            $      &$ K=100           $     \\
  \hline
  \gls{FA}             & $0.290\pm 0.003 $         &$ 0.275 \pm 0.003$     &$ 0.290 \pm 0.003 $      &$ 0.276 \pm 0.003 $     \\
  \gls{gemb} (c=10)    & $0.239\pm 0.006 $         &$ 0.239\pm 0.005 $     &$ 0.246 \pm 0.004 $      &$ 0.245 \pm 0.003 $     \\
  \gls{gemb} (c=50)    & $0.227\pm 0.002 $         &$ {\bf 0.222\pm 0.002}$&$ 0.235 \pm 0.003 $      &$ {\bf 0.232 \pm 0.003} $     \\
  \gls{ngemb} (c=10)   & $0.263\pm 0.004 $         &$ 0.261\pm 0.004 $     &$ 0.250 \pm 0.004 $      &$ 0.261 \pm 0.004 $     \\
  \bottomrule
\end{tabular}
\caption{Analysis of neural data: mean squared error and standard
  errors of neural activity (on the test set) for different models.
  Both \gls{efemb} models significantly outperform \gls{FA};
  \gls{gemb} is more accurate than \gls{ngemb}.
}
\label{tab:results}
\vspace{-7pt}
\end{table}
\begin{figure*}[t]
  \centering
  {\includegraphics[width=0.93\textwidth]{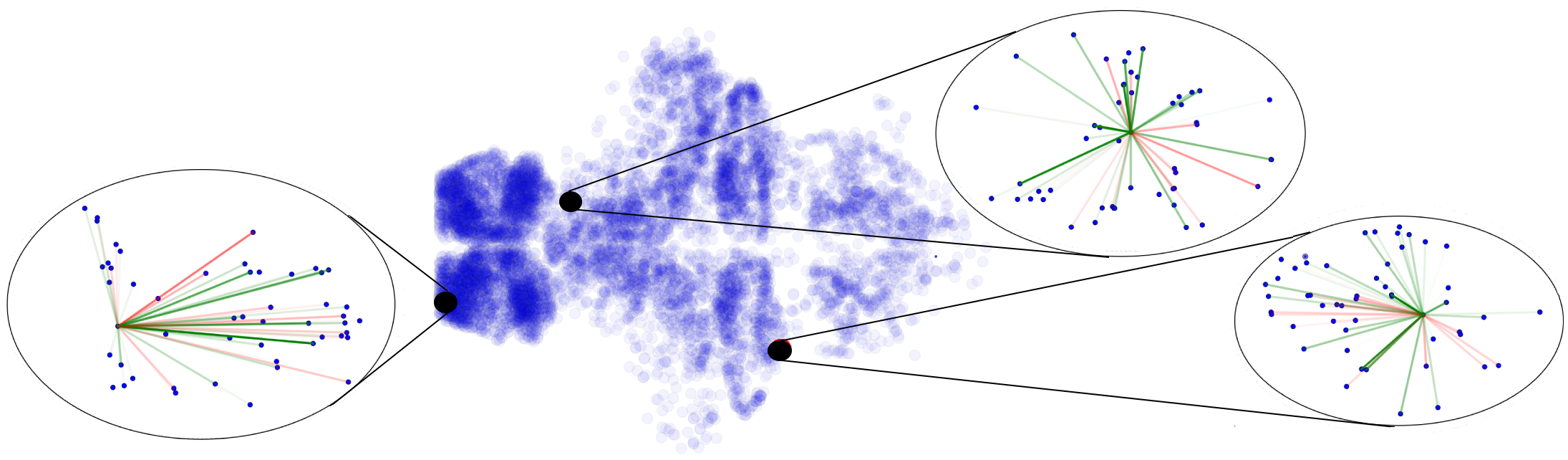}}
  \caption{ Top view of the zebrafish brain, with blue circles at the
    location of the individual neurons. We zoom on 3 neurons and their
    $50$ nearest neighbors (small blue dots), visualizing the
    ``synaptic weights'' learned by a \gls{gemb} model ($K=100$).
    The edge color encodes the inner product of the neural embedding
    vector and the context vectors $\rho_n^\top\alpha_m$ for
    each neighbor $m$. Positive values are green, negative values are
    red, and the transparency is proportional to the magnitude. With
    these weights we can hypothesize how nearby neurons interact. }
\label{fig:neuro}
\vspace{-5pt}
\end{figure*}

\parhead{Results.}
\ref{tab:results} reports both types of evaluation. The \gls{efemb}
models significantly outperform \gls{FA} in terms of mean squared
error on the test set. \gls{gemb} obtains the best results with $100$
components and a context size of $50$. \ref{fig:neuro} illustrates
how to use the learned embeddings to hypothesize connections between
nearby neurons.

\subsection{Count Data: Market Basket Analysis and Movie Ratings}
We study the Poisson models \gls{pemb} and \gls{apemb} on two
applications: shopping and movies.

\begin{table}[t]
\centering
\small
	\subfloat[Market basket analysis.\label{tab:results_iri_llh}]{
		\begin{tabular}{ccc} \toprule
			\textbf{Model} & $K=20$ & $K=100$ \\ \hline
			\acrshort{pemb}  	& $-7.497\pm 0.007$	& $-7.199\pm 0.008$ \\
			\acrshort{pemb} (dw)  	& $-7.110\pm 0.007$	& $\mathbf{-6.950\pm 0.007}$ \\
			\acrshort{apemb} 	& $-7.868\pm 0.005$	& $-8.414\pm 0.003$ \\
			\acrshort{HPF}  	& $-7.740\pm 0.008$	& $-7.626\pm 0.007$ \\
			Poisson \acrshort{PCA} 	& $-8.314\pm 0.009$	& $-11.01\pm 0.01$ \\ \bottomrule
		\end{tabular}%
	} \hspace*{5pt}
	\subfloat[Movie ratings.\label{tab:results_movielens_llh}]{
		\begin{tabular}{cc} \toprule
			 $K=20$ & $K=100$ \\ \hline
			 $\mathbf{-5.691\pm 0.006}$  & $-5.726\pm 0.005$ \\
                         $-5.790\pm 0.003$              & $-5.798\pm 0.003$ \\
			 $-5.964\pm 0.003$           & $-6.118\pm 0.002$ \\
			 $-5.787\pm 0.006$          	& $-5.859\pm 0.006$ \\
			 $-5.908\pm 0.006$           & $-7.50\pm 0.01$ \\ \bottomrule
		\end{tabular}
	}
\vspace{3pt}
\caption{\label{tab:results_poisson_llh}Comparison of predictive log-likelihood between \gls{pemb}, \gls{apemb}, \gls{HPF} \citep{Gopalan2015}, and Poisson \gls{PCA} \citep{collins2001generalization} on held out data. The \gls{pemb} model outperforms the matrix factorization models in both applications. For the shopping data, downweighting the zeros improves the performance of \gls{pemb}.}
\end{table}

\parhead{Market basket data.} We analyze the IRI dataset\footnote{We
  thank IRI for making
  the data available. All estimates and analysis in this paper, based
  on data provided by IRI, are by the authors and not by IRI.}
\citep{Bronnenberg2008}, which contains the purchases of anonymous
households in chain grocery and drug stores. It contains $137,632$
trips in 2012. We remove items that appear fewer than 10 times,
leaving a dataset with $7,903$ items. The context for each purchase is
the other purchases from the same trip.

\parhead{MovieLens data.} We also analyze the MovieLens-$100K$
dataset~\citep{harper2015movielens}, which contains movie ratings on a
scale from $1$ to $5$. We keep only positive ratings, defined to be
ratings of 3 or more (we subtract 2 from all ratings and set the negative ones to 0). 
The context of each rating is the other movies rated by the same user. After
removing users who rated fewer than $20$ movies
and movies that were rated fewer than $50$ times, the dataset contains
$777$ users and $516$ movies; the sparsity is about $5\%$.

\parhead{Models.} We fit the \gls{pemb} and the \gls{apemb} models
using number of components $K \in \{20, 100\}$. 
For each $K$ we select the Adagrad constant based on best predictive 
performance on the validation set. (The parameters we used are in \ref{tab:hyper}.)  
In these datasets, the distribution of the context size is heavy tailed. To handle
larger context sizes we pick a link function for the \gls{efemb}
model which rescales the sum over the context in \ref{eqn:linear-embedding} by the
context size (the number of terms in the sum).
We also fit a \gls{pemb} model that artificially downweights the contribution of
the zeros in the objective function by a factor of $0.1$, 
as done by \citet{hu2008} for matrix factorization.
We denote it as ``\gls{pemb} (dw).''

We compare the predictive performance with
\gls{HPF}~\citep{Gopalan2015} and Poisson
\gls{PCA}~\citep{collins2001generalization}. Both \gls{HPF} and
Poisson \gls{PCA} factorize the data into $K$-dimensional positive vectors of user
preferences, and $K$-dimensional positive vectors of item attributes.
\gls{apemb} and \gls{HPF} parameterize the mean additively; \gls{pemb}
and Poisson \gls{PCA} parameterize it multiplicatively. For the
\gls{efemb} models and Poisson \gls{PCA}, we use stochastic
optimization with $\ell_2$ regularization. For \gls{HPF}, we use
variational inference. See \ref{tab:hyper} in \ref{sec:hyper} for
details.

\parhead{Evaluation.} For the market basket data we hold out $5\%$ of
the trips to form the test set, also removing trips with fewer than
two purchased different items. In the MovieLens data we hold out
$20\%$ of the ratings and set aside an additional $5\%$ of the
non-zero entries from the test for validation.  We report prediction
performance based on the normalized log-likelihood on the test
set. For \gls{pemb} and \gls{apemb}, we compute the likelihood as the
Poisson mean of each nonnegative count (be it a purchase quantity or a
movie rating) divided by the sum of the Poisson means for all items,
given the context. To evaluate \gls{HPF} and Poisson \gls{PCA} at a
given test observation we recover the factor loadings using the other
test entries we condition on, and we use the factor loading to form
the prediction.

\parhead{Predictive performance.}
\ref{tab:results_poisson_llh} summarizes the test log-likelihood of
the four models, together with the standard errors across entries in
the test set. In both applications the \gls{pemb} model outperforms
\gls{HPF} and Poisson \gls{PCA}. On shopping data \gls{pemb} with
$K=100$ provides the best predictions; on MovieLens \gls{pemb} with
$K=20$ is best. For \gls{pemb} on shopping data, downweighting the
contribution of the zeros gives more accurate estimates.

\parhead{Item similarity in the shopping data.} Embedding models can
capture qualitative aspects of the data as well.
\ref{tab:iri_some_queries} shows four example products and their three
most similar items, where similarity is calculated as the cosine
distance between embedding vectors. (These vectors are from \gls{pemb}
with downweighted zeros and $K=100$.) For example, the most similar items to a soda are
other sodas; the most similar items to a yogurt are (mostly) other
yogurts.

\begin{table}
	\centering
	\scriptsize
	\begin{tabular}{llll} \toprule
		{\bf Maruchan chicken ramen}  		& {\bf Yoplait strawberry yogurt} 	& {\bf Mountain Dew soda}      & {\bf Dean Foods  1 \% milk} \\ \hline
		M. creamy chicken ramen		& Yoplait apricot mango yogurt        	& Mtn. Dew orange soda     & Dean Foods 2 \% milk \\
		M. oriental flavor ramen		& Yoplait strawberry orange smoothie   	& Mtn. Dew lemon lime soda & Dean Foods whole milk \\
		M. roast chicken ramen  		& Yoplait strawberry banana yogurt     	& Pepsi classic soda	       & Dean Foods chocolate milk \\ \bottomrule \\
	\end{tabular} 
	\caption{Top 3 similar items to a given example query words (bold face). The \gls{pemb} model successfuly captures similarities.\label{tab:iri_some_queries}}
\vspace{-10pt}
\end{table}

The \gls{pemb} model can also identify complementary and substitutable
products. To see this, we compute the inner products of the embedding
and the context vectors for all item pairs. A high value of the inner
product indicates that the probability of purchasing one item is
increased if the second item is in the shopping basket (i.e., they are
complements). A low value indicates the opposite effect and the items
might be substitutes for each other.

We find that items that tend to be purchased together have high value
of the inner product (e.g., potato chips and beer, potato chips and
frozen pizza, or two different types of soda), while items that are
substitutes have negative value (e.g., two different brands of pasta
sauce, similar snacks, or soups from different brands).  Other items
with negative value of the inner product are not substitutes, but they
are rarely purchased together (e.g., toast crunch and laundry
detergent, milk and a toothbrush).  \ref{sec:complements_substitutes}
gives examples of substitutes and complements.

\parhead{Topics in the movie embeddings.} The embeddings from
MovieLens data identify thematically similar movies. For each latent
dimension $k$, we sort the context vectors by the magnitude of the
$k$th component. This yields a ranking of movies for each component.
In \ref{sec:movie} we show two example rankings. (These are from a 
\gls{pemb} model with $K=50$.) The first one contains children's
movies; the second contains science-fiction/action movies.

\newpage
\section{Discussion}
\glsresetall

We described \glspl{efemb}, conditionally specified latent variable
models to extract distributed representations from high dimensional
data.  We showed that \gls{CBOW} \citep{mikolov2013distributed} is a
special case of \gls{efemb} and we provided examples beyond text: the
brain activity of zebrafish, shopping data, and movie ratings.  We fit
the \gls{efemb} objective usng stochastic gradients.  Our empirical
study demonstrates that an \gls{efemb} can better reconstruct data
than existing dimensionality-reduction techniques based on matrix
factorization.  Further, the learned embeddings capture interesting
semantic structure.

\subsection*{Acknowledgments}
This work is supported by the EU H2020 programme (Marie Sk\l{}odowska-Curie grant agreement 706760), NFS IIS-1247664, ONR N00014-11-1-0651, DARPA FA8750-14-2-0009, DARPA N66001-15-C-4032, Adobe, the John Templeton Foundation, and the Sloan Foundation.

\bibliographystyle{apa}
\small
\bibliography{myrefs}
\clearpage
\normalsize
\appendix
\glsresetall
\section*{\centering Supplement to Exponential Family Embeddings}
\section{Inference}
\label{sec:inference_long}

We fit the embeddings $\rho[i]$ and context vectors $\alpha[i]$ by
maximizing the objective function in \ref{eqn:objective}.  We use
\gls{SGD}.

We first calculate the gradient, using the identity for exponential
family distributions that the derivative of the log-normalizer is
equal to the expectation of the sufficient statistics, i.e.,
$\mathbb{E}[t(X)] = \nabla_{\eta} a(\eta)$.  With this result, the
gradient with respect to the embedding $\rho[j]$ is
\begin{align}
  \nabla_{\rho[j]}\mathcal{L}= \sum_{i=1}^I \big ( t(x_{i}) - \mathbb{E}[t(x_{i})]\big)\nabla_{\rho[j]}\eta_{i}+\nabla_{\rho[j]}\log p(\rho[j]).
  \label{eqn:gradient_embedding}
\end{align}
The gradient with respect to $\alpha[j]$ has the same form.  In
\ref{sec:sgd}, we detail this expression for the particular models
that we study empirically (\ref{sec:experiments}).

The gradient in \ref{eqn:gradient_embedding} can involve a sum of
many terms and be computationally expensive to compute. To alleviate
this, we follow noisy gradients using \gls{SGD}.  We form a subsample
$\mathcal{S}$ of the $I$ terms in the summation, i.e.,
\begin{align}
  \widehat{\nabla}_{\rho[j]} \mathcal{L} =
  \frac{I}{|\mathcal{S}|}\sum_{i\in\mathcal{S}}
  \big( t(x_{i}) - \mathbb{E}[t(x_{i})]\big)\nabla_{\rho[j]}\eta_{i} +
  \nabla_{\rho[j]} \log p(\rho[j]),
  \label{eqn:gradient_embedding_noisy}
\end{align}
where $|\mathcal{S}|$ denotes the size of the subsample 
and where we scaled the summation to ensure an unbiased estimator of the gradient.
\ref{eqn:gradient_embedding_noisy} reduces computational complexity
when $|\mathcal{S}|$ is much smaller than the total number of terms.
At each iteration of \gls{SGD} we compute noisy gradients with respect
to $\rho[j]$ and $\alpha[j]$ (for each $j$) and take gradient steps
according to a step-size schedule.  We use Adagrad~\citep{Duchi2011}
to set the step-size.

\parhead{Relation to negative sampling.} In language, particularly
when seen as a collection of binary variables, the data are sparse:
each word is one of a large vocabulary. When modeling sparse data, we
split the sum in \ref{eqn:gradient_embedding} into two contributions:
those where $x_{i}>0$ and those where $x_{i}=0$.  The gradient is
\begin{align}
  \nabla_{\rho[j]}\mathcal{L} = &
                                  \sum_{i: x_{i}>0}
                                  \big(t(x_{i})-\mathbb{E}[t(x_{i})]\big)
                                  \nabla_{\rho[j]}\eta_{i}
                                  +
                                  \sum_{i: x_{i}=0}
                                  \big(t(0) - \mathbb{E}[t(x_{i})]\big)
                                  \nabla_{\rho[j]}\eta_{i}
                                  \label{eqn:gradient_embedding_poisson} \\
                                & + \nabla_{\rho[j]} \log p(\rho[j]). \nonumber
\end{align}
We compute the first term of the gradient exactly---when the data is
sparse there are not many summations to make---and we estimate 
the second term with subsampling.  Compared to computing
the full gradient, this reduces the complexity when most of the
entries $x_{i}$ are zero.  But, it retains the strong information
about the gradient that comes from the non-zero entries.

This relates to negative sampling, which is used to approximate the
skip-gram objective~\citep{mikolov2013distributed}.  Negative sampling
re-defines the skip-gram objective to distinguish target (observed)
words from randomly drawn words, using logistic regression. The
gradient of the stochastic objective is identical to a noisy but
biased estimate of the gradient in
\ref{eqn:gradient_embedding_poisson} for a \gls{bemb} model.  To
obtain the equivalence, preserve the terms for the non-zero data and
subsample terms for the zero data.  While an unbiased stochastic
gradient would rescale the subsampled terms, negative sampling does
not.  It is thus a biased estimate, which down-weights the
contribution of the zeros.

\section{Stochastic Gradient Descent}
\label{sec:sgd}
To specify the gradients in Equation \ref{eqn:gradient_embedding} for the \gls{SGD} procedure  we need the sufficient statistic $t(x)$, the expected sufficient statistic $\mathbb{E}[t(x)]$, the gradient of the natural parameter with respect to the embedding vectors and the gradient of the regularizer on the embedding vectors. In this appendix we specify these quantities for the models we study empirically in Section \ref{sec:experiments}.
\subsection{Gradients for \gls{gemb}}
Using the notation $i = (n,t)$ and reflecting the embedding structure $\rho[i]=\rho_n$, $\alpha[i]=\alpha_n$, the gradients with respect to each embedding and each context vector becomes
 \label{sec:gaussian_gradients}
\begin{align}
\label{eqn:gaussian_gradient_rho}
\nabla_{\rho_n}\mathcal{L} &=- \lambda \rho_n + \frac{1}{\sigma^2} \sum_{t=1}^T \big(x_{(n,t)} - \rho_n^\top \sum_{m \in c_n} x_{(m,t)}\alpha_{m}\big)\big(\sum_{m \in c_n} x_{(m,t)}\alpha_{m}\big) \\
\label{eqn:gaussian_gradient_alpha}
\nabla_{\alpha_n}\mathcal{L} &=- \lambda \alpha_n + \frac{1}{\sigma^2} \sum_{t=1}^T \sum_{m |n \in c_m} \big(x_{(m,t)} - \rho_{m}^\top \sum_{r \in c_m} x_{(r,t)}\alpha_{r}\big)\big( x_{(n,t)}\rho_{m}\big) 
\end{align}
\subsection{Gradients for \gls{ngemb}}
 \label{sec:lognormal_gradients}
By restricting the parameters to be nonnegative we can learn nonnegative synaptic weights between neurons.
For notational simplicity we write the parameters as $\exp(\rho)$ and $\exp(\alpha)$ and update them in log-space. The operator $\circ$ stands for element wise multiplication. With this notation, the gradient for the \gls{ngemb} can be easily obtained from Equations \ref{eqn:gaussian_gradient_rho} and \ref{eqn:gaussian_gradient_alpha} by applying the chain rule.
\begin{align}
\label{eqn:nonnegative_gradient_rho}
\nabla_{\rho_n}\mathcal{L} =&- \lambda \exp(\rho_n)\circ \exp(\rho_n)\\
&+ \frac{1}{\sigma^2} \sum_{t=1}^T \big(x_{(n,t)} - \exp(\rho_n)^\top \sum_{m \in c_n} x_{(m,t)}\exp(\alpha_{m})\big)\big(\sum_{m \in c_n} x_{im}\exp(\rho_{n})\circ\exp(\alpha_{m}) \big)\nonumber \\
\label{eqn:nonnegative_gradient_alpha}
\nabla_{\alpha_n}\mathcal{L} =&- \lambda \exp(\alpha_{n})\circ\exp(\alpha_n) \\
&+ \frac{1}{\sigma^2} \sum_{t=1}^T \sum_{m |n \in c_m} \big(x_{(m,t)} - \exp(\rho_{m})^\top \sum_{r \in c_m} x_{(r,t)}\exp(\alpha_{r})\big)\big( x_{(n,t)}\exp(\rho_{m})\circ\exp(\alpha_{n})\big) \nonumber
\end{align}

\subsection{Gradients for \gls{pemb}}
\label{sec:poisson_gradients} 

We proceed similarly as for the \gls{gemb} model.

\begin{align}
\nabla_{\rho_n}\mathcal{L} &=- \lambda \rho_n + \sum_{t=1}^T \left(x_{(n,t)} - \exp\left(\rho_n^{\top} \sum_{m \in c_n} x_{(m,t)}\alpha_{m}\right)\right)\left(\sum_{m \in c_n} x_{(m,t)}\alpha_{m}\right) \\
\nabla_{\alpha_n}\mathcal{L} &=- \lambda \alpha_n + \sum_{t=1}^T \sum_{m |n \in c_m}  \left(x_{(m,t)}-\exp\left( \rho_{m}^{\top} \sum_{r \in c_m} x_{(r,t)}\alpha_{r} \right)\right) \left( x_{(n,t)}\rho_{m}\right)
\end{align}

\subsection{Gradients for \gls{apemb}}
\label{sec:additive_poisson_gradients} 

Here, we proceed in a similar manner as for the \gls{ngemb} model.

\begin{align}
\nabla_{\rho_n}\mathcal{L} &=- \lambda \exp(\rho_n)\circ \exp(\rho_n) + \sum_{t=1}^T \left(\frac{x_{(n,t)}}{\rho_n^{\top} \sum_{m \in c_n} x_{(m,t)}\alpha_{m}} - 1\right)\left(\sum_{m \in c_n} x_{(m,t)}\alpha_{m}\right) \\
\nabla_{\alpha_n}\mathcal{L} &=- \lambda \exp(\alpha_{n})\circ\exp(\alpha_n) + \sum_{t=1}^T \sum_{m |n \in c_m}  \left(\frac{x_{(m,t)}}{\rho_{m}^{\top} \sum_{r \in c_m} x_{(r,t)}\alpha_{r}}-1 \right) \left( x_{(n,t)}\rho_{m}\right)
\end{align}

\newpage
\section{Algorithm Details}
\label{sec:hyper}
\vspace{-15pt}
\begin{table}[h!]
\centering
\begin{tabular}{llllll}
\toprule
         &  {\bf Model}                   & \begin{tabular}[c]{@{}l@{}}minibatch\\ size\end{tabular} & \begin{tabular}[c]{@{}l@{}}regularization \\ parameter\end{tabular} & \begin{tabular}[c]{@{}l@{}}number\\iterations \end{tabular} & \begin{tabular}[c]{@{}l@{}}negative \\ samples\end{tabular}  \\
\hline
neuro    & \acrshort{gemb}        & $100$                                                    & $10$                                                                & $500$                                                      & n/a                                                            \\
neuro    & \acrshort{ngemb}       & $100$                                                    & $0.1$                                                                 & $500$                                                      & n/a                                                            \\
shopping & all models        & n/a                                                      & $1$                                                                & $3000$                                                      & $10$                                                           \\
movies   & all models          & n/a                                                      & $1$                                                                & $3000$                                                     & $10$                                         \\
\bottomrule
\end{tabular}
\caption{Algorithm details for the models studied in Section \ref{sec:experiments}.}
\label{tab:hyper}
\end{table}

\section{Complements and Substitutes in the Shopping Data}
\label{sec:complements_substitutes}
Table \ref{tab:iri_high_inner_prod} shows some pairs of items with high inner product of embedding vectors and context vector. The items in the first column have higher probability of being purchased if the item in the second column is in the shopping basket. We can observe that they correspond to items that are frequently purchased together (potato chips and beer, potato chips and frozen pizza, two different sodas).

Similarly, Table \ref{tab:iri_low_inner_prod} shows some pairs of items with low inner product. The items in the first column have lower probability of being purchased if the item in the second column is in the shopping basket. We can observe that they correspond to items that are rarely purchased together (detergent and toast crunch, milk and toothbrush), or that are substitutes of each other (two different brands of snacks, soup, or pasta sauce).
\vspace{-5pt}
\begin{table}[h]
	\centering
  \scriptsize
	\begin{tabular}{lll}\toprule
		Inner product & Item 1 & Item 2 \\ \hline
    $2.12$ & Diet 7 Up lemon lime soda & Diet Squirt citrus soda \\
    $2.11$ & Old Dutch original potato chips & Budweiser Select 55 Lager beer \\
    $2.00$ & Lays potato chips & DiGiorno frozen pizza \\
    $2.00$ & Coca Cola zero soda & Coca Cola soda \\
    $1.99$ & Soyfield vanilla organic yogurt & La Yogurt low fat mango \\
		\bottomrule
	\end{tabular}
	\caption{Market basket: List of several of the items with high inner product values. Items from the first column have higher probability of being purchased when the item in the second column is in the shopping basket.\label{tab:iri_high_inner_prod}}
\end{table}
\vspace{-25pt}
\begin{table}[h]
	\centering
  \scriptsize
	\begin{tabular}{lll}\toprule
		Inner product & Item 1 & Item 2 \\ \hline
    $-5.06$ & General Mills cinnamon toast crunch & Tide Plus liquid laundry detergent \\
    $-5.00$ & Doritos chilli pepper & Utz cheese balls \\
    $-5.00$ & Land O Lakes 2\% milk & Toothbrush soft adult (private brand) \\
    $-5.00$ & Beef Swanson Broth soup 48oz & Campbell Soup cans 10.75oz \\
    $-4.99$ & Ragu Robusto saut\'{e}ed onion \& garlic pasta sauce & Prego tomato Italian pasta sauce \\
    \bottomrule
	\end{tabular}
	\caption{Market basket: List of several of the items with low inner product values. Items from the first column have lower probability of being purchased when the item in the second column is in the shopping basket.\label{tab:iri_low_inner_prod}}
\end{table}

\newpage
\section{Movie Rating Results}
\label{sec:movie}
\ref{tab:kids_movies,tab:scifi_movies} show clusters of ranked movies that are learned by our \gls{pemb} model.
These rankings were generated as follows. For each latent dimension $k\in\{1,\cdots,K\}$ we sorted the context vectors according their value in this dimension.
This gives us a ranking of context vectors for every $k$. 
\ref{tab:kids_movies,tab:scifi_movies} show the 10 top items of the ranking for two different values of $k$. Similar as in topic modeling, the latent dimensions have the interpretation of topics. We see that sorting the context vectors this way reveals thematic structure in the collection of movies. While Table \ref{tab:kids_movies} gives a table of movies for children, Table \ref{tab:scifi_movies} shows a cluster of science-fiction and action movies (with a few outliers). 
\begin{table}[h]
\centering
\begin{tabular}{llll}\toprule
	\# & Movie Name                         & Year &  Rank       \\ \hline
1 & Winnie the Pooh and the Blustery Day & 1968 & $0.62$ \\
2 & Cinderella & 1950 & $0.50$ \\
3 & Toy Story & 1995 & $0.46$ \\
4 & Fantasia & 1940 & $0.44$ \\
5 & Dumbo & 1941 & $0.43$ \\
6 & The Nightmare Before Christmas & 1993 & $0.37$ \\
7 & Snow White and the Seven Dwarfs & 1937 & $0.37$ \\
8 & Alice in Wonderland & 1951 & $0.35$ \\
9 & James and the Giant Peach & 1996 & $0.35$ \\ \bottomrule
\end{tabular}
\caption{\label{tab:kids_movies} Movielens: Cluster for ``kids movies''.}
\end{table}
\begin{table}[h]
\centering
\begin{tabular}{llll}\toprule
\# & Movie Name                         & Year &  Rank       \\ \hline
1 & Die Hard: With a Vengeance & 1995 & $1.25$ \\
2 & Stargate & 1994 & $1.19$ \\
3 & Star Trek IV: The Voyage Home & 1986 & $1.14$ \\
4 & Manon of the Spring (Manon des sources) & 1986 & $1.14$ \\
5 & Fifth Element, The & 1997 & $1.14$ \\
6 & Star Trek VI: The Undiscovered Country & 1991 & $1.13$ \\
7 & Under Siege & 1992 & $1.11$ \\
8 & GoldenEye & 1995 & $1.07$ \\
9 & Supercop & 1992 & $1.07$ \\ \bottomrule
\end{tabular}
\caption{\label{tab:scifi_movies} Movielens: Cluster for ``science-fiction/action movies''.}
\end{table}

\end{document}